\documentclass[runningheads]{llncs}

\usepackage{tikz}
\usepackage{comment}
\usepackage{amsmath,amssymb} 
\usepackage{color}

\usepackage{soul}
\usepackage{url}
\usepackage[hidelinks]{hyperref}
\usepackage[utf8]{inputenc}
\usepackage[small]{caption}
\usepackage{graphicx}
\usepackage{booktabs}
\usepackage{algorithm}
\usepackage{color}
\usepackage{wrapfig}

 \usepackage{amsfonts}
 \usepackage{algpseudocode} 
 \usepackage{bm}
 \usepackage{makecell} 
 \usepackage{amsmath,subfigure,epsfig,bbding}
 \usepackage[tableposition=top]{caption}
 \usepackage[figuresright]{rotating}
 \usepackage{multirow}
 \usepackage{booktabs} 
 \usepackage{array, boldline, rotating} 
 \usepackage{relsize}
 
 \newcolumntype{P}[1]{>{\centering\arraybackslash}p{#1}}
 \makeatother

\usepackage{array}
\makeatletter
\newcommand{\thickhline}{%
	\noalign {\ifnum 0=`}\fi \hrule height 1pt
	\futurelet \reserved@a \@xhline
}
\newcolumntype{"}{@{\vrule width 1pt}}
\makeatother

\begin{document}
	\pagestyle{headings}
	\mainmatter
	\def\ECCVSubNumber{7}  
	
	\title{See Finer, See More: Implicit Modality Alignment for Text-based Person Retrieval} 
	
	\titlerunning{See Finer, See More}
	
	\author{Xiujun Shu\inst{1}\thanks{Equal contribution \qquad   $^\dagger$Corresponding Author}, Wei Wen\inst{1}\printfnsymbol{1}, Haoqian Wu\inst{1}, Keyu Chen\inst{1}, Yiran Song\inst{1,2}, \\Ruizhi Qiao\inst{1}, Bo Ren\inst{1}, Xiao Wang\inst{3}$^\dagger$}
	\authorrunning{Xiujun Shu, Wei Wen et al.}
	\institute{$^1$Tencent Youtu Lab \\ \email{\{xiujunshu, jawnrwen, linuswu, yolochen, ruizhiqiao, timren\}@tencent.com} \quad 
		$^2$Shanghai Jiao Tong University \quad $^3$Anhui University  \\ \email{songyiran@sjtu.edu.cn, wangxiaocvpr@foxmail.com}
	}

	\makeatletter
	\newcommand{\printfnsymbol}[1]{%
		\textsuperscript{\@fnsymbol{#1}}%
	}
	\makeatother

	\maketitle
	
	\begin{abstract}
		Text-based person retrieval aims to find the query person based on a textual description. The key is to learn a common latent space mapping between visual-textual modalities. To achieve this goal, existing works employ segmentation to obtain explicitly cross-modal alignments or utilize attention to explore salient alignments.  
		These methods have two shortcomings: 1) Labeling cross-modal alignments are time-consuming. 2) Attention methods can explore salient cross-modal alignments but may ignore some subtle and valuable pairs. To relieve these issues, we introduce an \textbf{I}mplicit \textbf{V}isual-\textbf{T}extual (\textbf{IVT}) framework for text-based person retrieval. Different from previous models, IVT utilizes a single network to learn representation for both modalities, which contributes to the visual-textual interaction. To explore the fine-grained alignment, we further propose two implicit semantic alignment paradigms: multi-level alignment (MLA) and bidirectional mask modeling (BMM). The MLA module explores \textbf{finer} matching at sentence, phrase, and word levels, while the BMM module aims to mine \textbf{more} semantic alignments between visual and textual modalities. Extensive experiments are carried out to evaluate the proposed IVT on public datasets, \emph{i.e.,} CUHK-PEDES, RSTPReID, and ICFG-PEDES. Even without explicit body part alignment, our approach still achieves state-of-the-art performance. Code is available at: \textcolor{blue}{\url{https://github.com/TencentYoutuResearch/PersonRetrieval-IVT}}.

		\keywords{Text-based Person Retrieval, Person Search by Language, Cross-Model Retrieval}
	\end{abstract}

	\section{Introduction}
	Person re-identification (re-ID) has many applications, \emph{e.g.,} finding suspects or lost children in surveillance, and tracking customers in supermarkets. As a sub-task of person re-ID, text-based person retrieval (TPR) has attracted remarkable attention in recent years~\cite{Shuang2017Person,zheng2020hierarchical,wang2021text}. This is due to the fact that textual descriptions are easily accessible and can describe more details in a natural way. For example, police officers usually access surveillance videos and take the deposition from witnesses. Textual descriptions can provide complementary information and even are critical in scenes where images are missing.
	
	\begin{figure}[t]
		\centering
		\includegraphics[width=\linewidth]{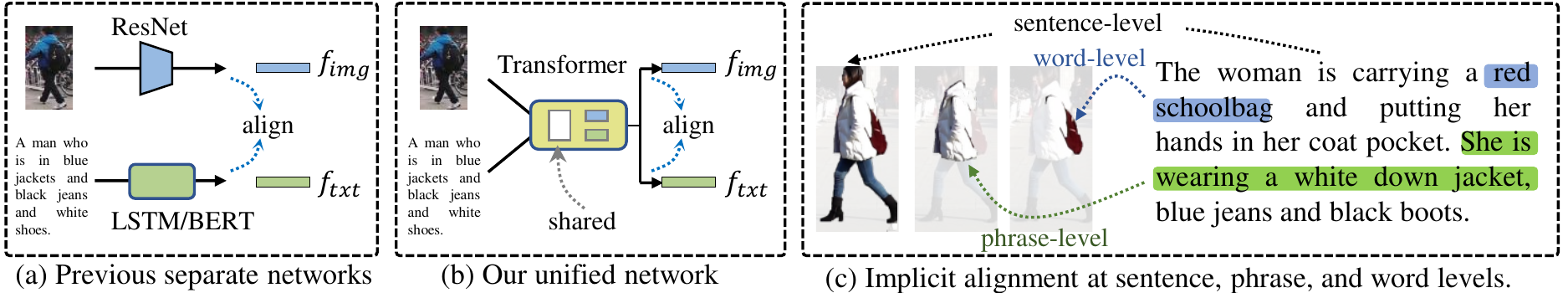} 
		\caption{\textbf{Illustration of the key idea of IVT.} 
			Previous methods use separate models to extract features, while we utilize a single network for both modalities. The shared parameters contribute to learning a common space mapping. Besides, we explore semantic alignment using three-level matchings. Not only see finer, but also see more semantic alignments.
		} 
		\label{fig:motivation} 
	\end{figure}

	Text-based person retrieval needs to process visual and textual modalities, and its core is to learn a common latent space mapping between them. To achieve this goal, current works~\cite{wang2020vitaa,jing2020pose} firstly utilize different models to extract features, \emph{i.e.,} ResNet50~\cite{he2016deep} for visual modality, and LSTM ~\cite{zhang2015bidirectional} or BERT~\cite{kenton2019bert} for textual modality. Then they are devoted to exploring visual-textual part pairs for semantic alignment. However, these methods have at least two drawbacks that may lead to suboptimal cross-modal matching. \textbf{First}, separate models lack modality interaction. Each model usually contains many layers with a large number of parameters, and it is difficult to achieve full interaction just using a matching loss at the end. To relieve this issue, some works~\cite{zhang2021vinvl,li2021align} on general image-text pre-training use cross-attention to conduct interaction. However, they require to encode all possible image-text pairs to compute similarity scores, which leads to quadratic time complexity at the inference stage. How to design a more suitable network for the TPR task still needs profound thinking. \textbf{Second}, labeling visual-textual part pairs, \emph{e.g.,} head, upper body, and lower body, is time-consuming, and some pairs may be missing due to the variability of textual descriptions. For example, some text contains descriptions of hairstyles and pants, but others do not contain this information. Some researchers begin to explore implicit local alignment to mine part matching \cite{wang2021text,zheng2020hierarchical}. To ensure reliability, partial local matchings with high confidence are usually selected. However, these parts usually belong to salient regions that can be easily mined by global alignment, \emph{i.e.,} they do not bring additional information gain. According to our observation, local semantic matching should not only see \textbf{finer}, but also see \textbf{more}. Some subtle visual-textual cues, \emph{e.g.,} hairstyle and logo on clothes, maybe easily ignored but could be complementary to global matching.

	To solve the above problems, we first introduce an \textbf{I}mplicit \textbf{V}isual-\textbf{T}extual (\textbf{IVT}) framework, which can learn representation for both modalities only using a single network (See Fig.~\ref{fig:motivation}(b)). This benefits from the merit that Transformer can operate on any modality that can be tokenized. To avoid the shortcomings of separate models and cross-attention, \emph{i.e.,} separate models lack modality interaction and cross-attention models are quite slow at the inference stage, IVT supports separate feature extracting to ensure retrieval speed and shares some parameters that contribute to learning a common latent space mapping. To explore fine-grained modality matching, we further propose two implicit semantic alignment paradigms: multi-level alignment (MLA) and bidirectional mask modeling (BMM). The two paradigms do not require extra manual labeling and can be easily implemented. Specifically, as shown in Fig.~\ref{fig:motivation}(c), MLA aims to explore fine-grained alignment by using sentence, phrase, and word-level matchings. BMM shares a similar idea with MAE~\cite{he2021masked} and BEIT~\cite{bao2021beit} in that both learn better representation through random masking. The difference is that the latter two aim at single-modal autoencoding-style reconstruction, while BMM does not reconstruct images but focuses on learning cross-modal matching. By masking a certain percentage of visual and textual tokens, BMM forces the model to mine more useful matching cues. The proposed two paradigms could not only see finer but also see more semantic alignments. Extensive experiments demonstrate the effectiveness on the TPR task.

	Our contributions can be summarized as three folds: (1) We propose to tackle the modality alignment from the perspective of backbone network and introduce an Implicit Visual-Textual (IVT) framework. This is the first unified framework for text-based person retrieval. (2) We propose two implicit semantic alignment paradigms, \emph{i.e.,} MLA and BMM, which enable the model to mine finer and more semantic alignments. (3) We conduct extensive experiments and analyses on three public datasets. Experimental results show that our approach achieves state-of-the-art performance.

	\section{Related Work} 
	\noindent 
	\textbf{Text-based Person Retrieval. } 
	Considering the great potential economic and social value of text-based person retrieval, Li et al. propose the first benchmark dataset CUHK-PEDES \cite{Shuang2017Person} in 2017, and also build a baseline, \emph{i.e.,} GNA-RNN, based on LSTM network. Early works utilize ResNet50 and LSTM to learn representations for visual-textual modalities, and then utilize matching loss to align them. For example, CMPM~\cite{zhang2018deep}  associates the representations across different modalities using KL divergence. Besides aligning the features, some works study the identity cue~\cite{li2017identity}, which helps learn discriminative representations. Since text-based person retrieval requires fine-grained recognition of human bodies, later works start to explore global and local associations. Some works~\cite{chen2018,chen2018improving} utilize visual-textual similarity to mine part alignments. ViTAA \cite{wang2020vitaa} segments the human body and utilizes k-reciprocal sampling to associate the visual and textual attributes. Surbhi et al. \cite{aggarwal2020text} propose to create semantic-preserving embeddings through attribute prediction. Since visual and textual attributes require pre-processing, more works attempt to use attention mechanisms to explore fine-grained alignment. PMA~\cite{jing2020pose} proposes a pose-guided multi-granularity attention network. 
	HGAN~\cite{zheng2020hierarchical} splits the images into multi-scale stripes and utilizes attention to select top-k part alignments. Other works include adversarial learning and relation modeling. TIMAM~\cite{ARL} learns modality-invariant feature representations using adversarial and cross-modal matching objectives. A-GANet~\cite{liu2019deep} introduces the textual and visual scene graphs consisting of object properties and relationships. In summary, most current works learn modality alignment by exploiting local alignments. In this work, we study the modality alignment from different perspectives, in particular, how to obtain full modality interaction and how to achieve local alignment simply. The proposed framework can effectively address these issues and achieve satisfying performance.

	\noindent \textbf{Transformer and Image-Text Pre-Training. }
	Transformer \cite{vaswani2017Transformer} is firstly proposed for machine translation in the natural language processing (NLP) community. After that, many follow-up works are proposed and set new state-of-the-art one after another, such as BERT \cite{kenton2019bert}, GPT series \cite{radford2018GPT,radford2019GPT2,brown2020GPT3}. The research on Transformer-based representations in computer vision is also becoming a hot spot. Early works like ViT \cite{dosovitskiy2020image} and Swin Transformer \cite{liu2021swinTransformer} take the dividing patches as input, like the discrete tokens in NLP, and achieve state-of-the-art performance on many downstream tasks. Benefiting from the merit that Transformer can operate on any modality that can be tokenized, it has been utilized in the multi- or cross-modal tasks intuitively~\cite{wang2022MMPTMSurvey}. Lu et al. \cite{lu2019vilBERT} propose the ViLBERT to process both visual and textual inputs in separate streams that interact through co-attentional Transformer layers. Oscar~\cite{li2020oscar} and VinVL~\cite{zhang2021vinvl} take the image, text, and category tags as inputs and find that the category information and stronger object detector can bring better results. Many recent works study which architecture is better for multi- or cross-modal tasks, \emph{e.g.,} UNITER~\cite{chen2020uniter}, ALBEF~\cite{li2021align}, and ViLT~\cite{kim2021vilt}. These works generally employ several training objectives to support multiple downstream vision-language tasks. The most relevant downstream task for us is image-text retrieval. To obtain modality interaction, these works generally employ cross-attention in the fusion blocks. However, they have a very slow retrieval speed at the inference stage because they need to predict the similarity of all possible image-text pairs. The ALIGN~\cite{jia2021ALIGN} and CLIP~\cite{radford2021CLIP} are large-scale vision-language models pre-trained using only contrastive matching. These separate models are suitable for image-text retrieval, but they generally achieve satisfying performance in zero-shot settings. Some works, \emph{e.g.,} switch Transformers~\cite{fedus2021switch}, VLMO~\cite{wang2021vlmo}, attempt to optimize the network structure so that both retrieval and other visual-language tasks can be supported. This paper is greatly inspired by these works and aims to relieve the modality alignment for text-based person retrieval.

	\begin{figure*}[t]
		\centering  
		\includegraphics[width=11cm]{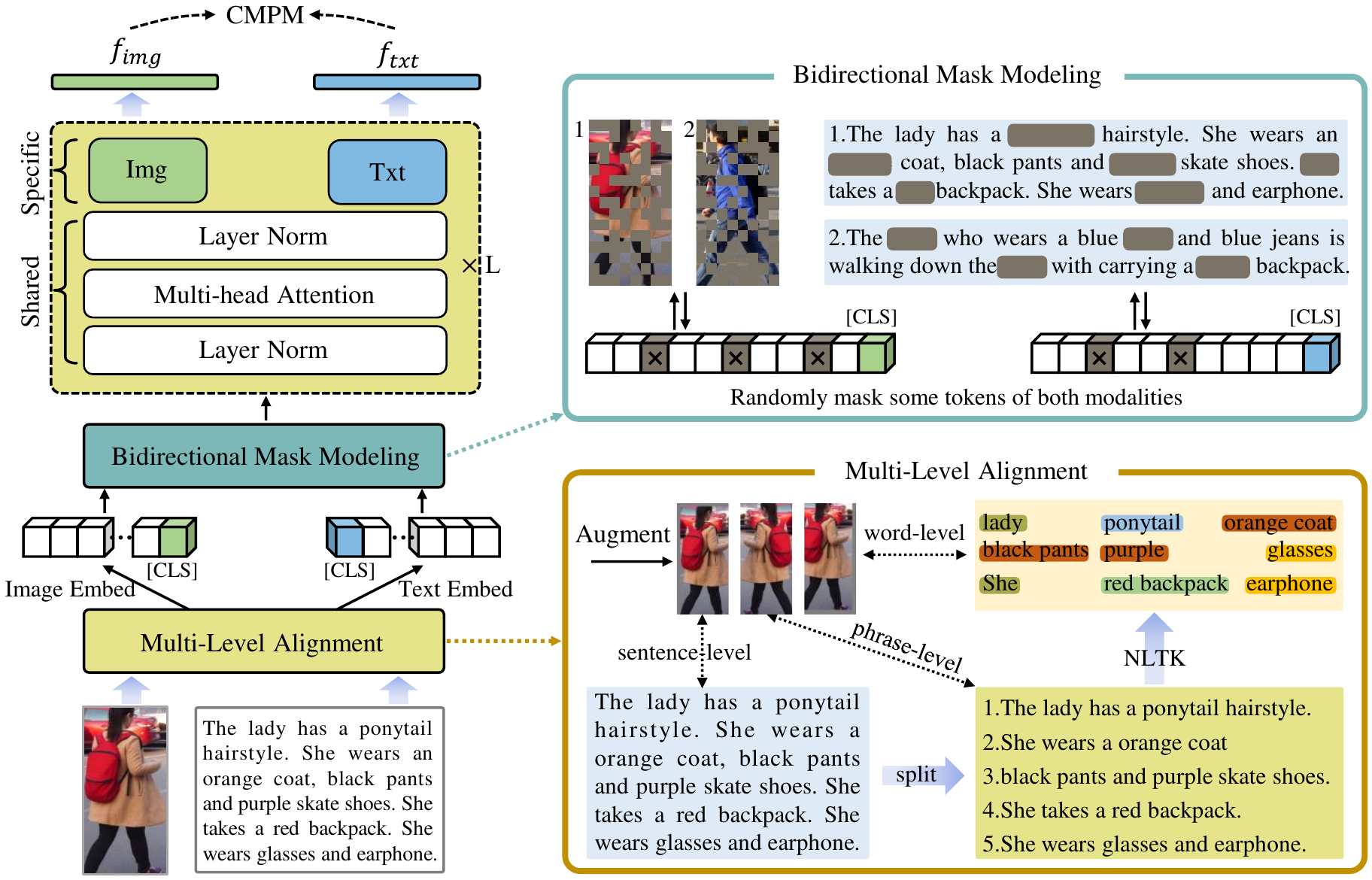} 
		\caption{\textbf{Architecture of the proposed IVT Framework.} 
			It consists of a unified visual-textual network and two implicit semantic alignment paradigms, \emph{i.e.,} multi-level alignment (MLA) and bidirectional mask modeling (BMM). The unified visual-textual network contains parameter-shared and specific modules, which contribute to learning common space mapping. MLA aims to see ``\textcolor{blue}{finer}'' by exploring local and global alignments from three-level matchings. BMM aims to see ``\textcolor{blue}{more}'' by mining more semantic alignments from random masking for both modalities.}	
		\label{fig:framework} 
	\end{figure*}

	
	\section{METHODOLOGY}
	
	\subsection{Overview} 
	To tackle the modality alignment in text-based person retrieval, we propose an Implicit Visual-Textual (IVT) framework as shown in Fig.~\ref{fig:framework}. It consists of a unified visual-textual network and two implicit semantic alignment paradigms, \emph{i.e.,} multi-level alignment (MLA) and bidirectional mask modeling (BMM). One key idea of IVT lies in tackling the modality alignment using a unified network. By sharing some modules, \emph{e.g.,} the layer normalization and multi-head attention, the unified network contributes to learning common space mapping between visual and textual modalities. It can also learn modality-specific cues using different modules. The two implicit semantic alignment paradigms, \emph{i.e.,} MLA and BMM, are proposed to explore fine-grained alignment. Different from previous methods that use manually processed parts or select salient parts from attention, the two paradigms could mine not only finer but also more semantic alignments, which is another key idea of our proposed IVT.

	\subsection{Unified Visual-Textual Network}

	\subsubsection{Embedding.}
	As shown in Fig.~\ref{fig:framework}, the input are image-text pairs, which provide the appearance characteristics of a person from visual and textual modalities. Let the image-text pairs denoted as $\{x_i,t_i,y_i\}|_{i=1}^N$, where $x_i,t_i,y_i$ denote the image, text, and identity label, respectively. $N$ is the total number of samples. For an input image $x_i\in \mathbb{R}^{H\times W\times C}$, it is firstly split into $K=H\cdot W/P^2$ patches, where $P$ denotes the patch size, and then linearly projected into patch embeddings $\{f_k^v\}|_{k=1}^K$. This operation can be realized using a single convolutional layer. The patch embeddings are then prepended with a learnable class token $f_{cls}^v$, and added with a learnable position embedding $f_{pos}^v$ and a type embedding $f_{type}^v$.
	\begin{equation}
	\textbf{f}^v=[f_{cls}^v,f_1^v,...,f_K^v]+f_{pos}^v+f_{type}^v.
	\end{equation} 
	
	For the input text $t_i$, it generally consists of one or several sentences and each sentence has a sequence of words. The pretrained word embedding is leveraged to project words into token vectors:
	\begin{equation}
	\textbf{f}^t=[f_{cls}^t,f_1^t,...,f_M^t,f_{sep}^t]+f_{pos}^t+f_{type}^t,
	\end{equation} 
	where $f_{cls}^t$ and $f_{sep}^t$ denote the start and end tokens. $M$ indicates the length of tokenized subword units. $f_{pos}^t$ is the position embedding and $f_{type}^t$ is the type embedding.

	\subsubsection{Visual-Textual Encoder.}
	Current works on the TPR task utilize separate models, which lack full modality interaction. Some recent works on general image-text pre-training attempt to utilize cross-attention to achieve modality interaction. However, cross-attention requires encoding all possible image-text pairs at the inference stage, which leads to a very slow retrieval speed. Based on these observations, we propose to take the unified visual-textual network for the TPR task. The unified network has a quick retrieval speed and supports modality interaction.

	As shown in Fig.~\ref{fig:framework}, the network follows a standard architecture of ViT~\cite{dosovitskiy2020image} and stacks $L$ blocks in total. In each block, two modalities share layer normalization (LN) and multi-head self-attention (MSA), which contribute to learning common space mapping between visual and textual modalities. It is because the shared parameters help learn common data statistics. For example, LN would calculate the mean and standard deviation of input token embeddings, and shared LN would learn the statistically common values of both modalities. This can be regarded as a ``modality interaction'' from a data-level perspective. As the visual and textual modalities are not the same, each block has modality-specific feed-forward layers, \emph{i.e.,} the ``Img'' and ``Txt'' modules in Fig.~\ref{fig:framework}. They are used to capture modality-specific information by switching to visual or textual inputs. The complete processing in each block can be denoted as follows:
	\begin{align}
	\textbf{f}_i^{v/t}&={\rm MSA}({\rm LN}(\textbf{f}_{i-1}^{v/t}))+\textbf{f}_{i-1}^{v/t},\\ 
	\textbf{f}_i^{v/t}&={\rm MLP_{img/txt}}({\rm LN}(\textbf{f}_i^{v/t}))+\textbf{f}_i^{v/t},  
	\end{align}
	where ${\rm LN}$ denotes the layer normalization and ${\rm MSA}$ denotes the multi-head attention. $i$ is the index of the blocks. $\textbf{f}_{i-1}^{v/t}$ is the visual or textual output of the $(i-1)^{th}$ block and also the input of the $i^{th}$ block. ${\rm MLP_{img/txt}}$ denotes the modality-specific feed-forward layers. $\textbf{f}_i^{v/t}$ is the output of the $i^{th}$ block.

	\subsubsection{Output.} 
	The class token of the last block serves as the global representation, \emph{i.e.,} $f_{img}$ and $f_{txt}$ in Fig.~\ref{fig:framework}. 
	The dimension of both feature vectors are 768, and the outputs are normalized using the LN layer.  
	
	\subsection{Implicit Semantic Alignment}
	\subsubsection{Multi-Level Alignment}  
	Fine-grained alignment has been demonstrated to be the key to achieving performance improvement, such as segmented attributes~\cite{wang2020vitaa} and stripe-based parts~\cite{gao2021contextual,zheng2020hierarchical}.  
	These methods can be regarded as explicit part alignment, namely telling the model which visual-textual parts should be aligned. In this work, we propose an implicit alignment method, \emph{i.e.,} multi-level alignment, which is intuitive but very effective.

	As shown in Fig.~\ref{fig:framework}, the input image is firstly augmented to get three types of augmented ones, \emph{e.g.,} horizontal flipping and random cropping.  
	The input text generally consists of one or several sentences. We split them into more short sentences according to periods and commas. These short sentences are regarded as ``phrase-level'' representation, which describe partial appearance characteristics of the human body. To mine finer parts, we further utilize the  Natural Language Toolkit (NLTK)~\cite{loper2002nltk} to extract nouns and adjectives, which describe specific local characteristics, \emph{e.g.,} bag, clothes, or pants. The three-level textual descriptions, \emph{i.e.,} sentence-level, phrase-level, and word-level, correspond to the three types of augmented images. The three-level image-text pairs are randomly generated at each iteration. In this way, we construct a matching process that gradually refines from global to local, forcing the model to mine finer semantic alignments.

	The major difference between our approach and previous work is that we do not explicitly define visual semantic parts, instead, automatically explore aligned visual parts guided by three-level textual descriptions.
	This is due to the following observations that inspire our TPR framework:
	1) Previous explicit aligned methods lack inconsistency between the training and inference phases. During training, previous methods utilize an unsupervised way to explore local alignments, e.g., select the top-k salient alignments based on similarities. During the inference stage, only the global embeddings for each modality would be used, resulting in the inconsistent issue. 
	2) The explicit local alignment makes the training easier but makes it harder at the inference stage. Oversimpilified task design leads to worse generalization performance of the model.
	Even though we do not provide visual parts, but rather the full images, the model tries hard to mine local alignment at the training stage, thus achieving better performance at the inference stage due to the consistency.

	\subsubsection{Bidirectional Mask Modeling} 
	To automatically mine local alignment, recent methods~\cite{gao2021contextual,zheng2020hierarchical} split images into stripes and utilize attention to select top-k part alignments. However, they ignore the fact that top-k part alignments are usually salient cues, which may have been mined by the global alignment. Therefore, these parts may bring limited information gains. We argue that local alignments should not only be finer, but also more diverse. Some subtle visual-textual cues may be complementary to global alignment.
	
	As shown in Fig.~\ref{fig:framework}, we propose a bidirectional mask modeling (BMM) method to mine more semantic alignments. For the image and text tokens, we randomly mask some percentage of them and then force the visual and textual outputs to keep in alignment. In general, the masked tokens correspond to specific patches of the image or words of the text. If specific patches or words are masked, the model would try to mine useful alignment cues from other patches or words. Let us take the lady in Fig.~\ref{fig:framework} as an example, if the salient words ``orange coat'' and ``black pants'' are masked, the model would pay more attention to other words, \emph{e.g.,} ponytail hairstyle, red backpack. In this way, more subtle visual-textual alignments can be explored. At the training stage, this method makes it more difficult for the model to align image and text but helps it to mine more semantic alignments at the inference stage.
	
	The above method shares a similar idea to Random Erasing~\cite{zhong2020random}, MAE~\cite{he2021masked}, and BEIT~\cite{bao2021beit}. However, Random Erasing masks only one region. MAE and BEIT aim at image reconstruction using an autoencoding style. The proposed BMM does not reconstruct images but focuses on cross-modal matching.

	\subsection{Loss Function}
	We utilize the commonly used cross-modal projection matching (CMPM) loss~\cite{zhang2018deep} to learn visual-textual alignment, which is defined as follows:  
	\begin{align}  
	\small 
	&\mathcal{L}_{cmpm} = \frac{1}{B}\sum_{i=1}^{B}\sum_{j=1}^{B}\Big(p_{i,j}\cdot\log \frac{p_{i,j}}{q_{i,j}+ \epsilon}\Big), \\
	&p_{i,j} = \frac{exp(f_i^T\cdot f_j)}{\sum_{k=1}^{B}exp(f_i^T\cdot f_k)}, ~~~~~~~~ q_{i,j} = \frac{y_{i,j}}{\sum_{k=1}^{B}y_{i,k}}, 
	\end{align} 
	where $p_{i,j}$ denotes the matching probability. $f_i$ and $f_j$ denote the global features of different modalities. $B$ is the mini-batch size. $q_{i,j}$ denotes the normalized true matching probability. $\epsilon$ is a small number to avoid numerical problems.
	
	The CMPM loss represents the KL divergence from distribution $\textbf{q}$ to $\textbf{p}$. Following previous work~\cite{zhang2018deep}, the matching loss is computed in two directions, \emph{i.e.,} image-to-text and text-to-image. The total loss can be denoted as follows:
	\begin{equation}  
	\mathcal{L} = \mathcal{L}_{cmpm}^{t2v} + \mathcal{L}_{cmpm}^{v2t}. 
	\end{equation}


	\begin{figure}
		\centering
		\begin{minipage}[t]{0.45\linewidth}
			\centering
			\small
			\makeatletter\def\@captype{table}\makeatother\caption{Comparison with SOTA methods on CUHK-PEDES.}
			\centering
			\label{tab:sota_CUHK}
			\resizebox{1\linewidth}{!}{ 
				\renewcommand\arraystretch{1} 
				\begin{tabular}[h!]{p{2.6cm}|m{1.2cm}<{\centering}m{1.2cm}<{\centering}m{1.2cm}<{\centering}}
					\thickhline
					\textbf{Method} & \textbf{R1} & \textbf{R5} & \textbf{R10}  \\
					\hline
					CNN-RNN \cite{reed2016learning} &	 8.07  & -  & 32.47 \\ 
					GNA-RNN \cite{Shuang2017Person} & 19.05  & -  & 53.64 \\ 
					PWM-ATH \cite{chen2018} & 27.14  & 49.45  & 61.02 \\ 
					GLA \cite{chen2018improving} & 43.58  & 66.93  & 76.26 \\
					MIA~\cite{niu2020improving} &	53.10 & 75.00 & 82.90 \\
					A-GANet \cite{liu2019deep} & 53.14 & 74.03 & 81.95 \\  
					ViTAA \cite{wang2020vitaa} & 55.97 & 75.84 & 83.52 \\ 
					IMG-Net \cite{wang2020img} & 56.48 & 76.89 & 85.01 \\
					CMAAM \cite{aggarwal2020text} &	 56.68 &	77.18 &	84.86 \\
					HGAN \cite{zheng2020hierarchical} &	59.00 & 79.49 & 86.60 \\
					NAFS \cite{gao2021contextual} &	59.94 & 79.86 & 86.70 \\
					DSSL \cite{zhu2021dssl} & 59.98 & 80.41 & 87.56 \\
					MGEL~\cite{wang2021text}        &60.27  &80.01  &86.74 \\
					SSAN~\cite{ding2021semantically}   &61.37  &80.15  &86.73 \\
					NAFS~\cite{gao2021contextual}   &61.50  &81.19  &87.51 \\
					TBPS~\cite{han2021text}         &61.65  &80.98  &86.78 \\ 
					TIPCB~\cite{chen2022tipcb}      &63.63  &82.82  &89.01 \\ 
					\hline
					Baseline (Ours) &	55.75 & 75.68 &	84.13 \\
					\textbf{IVT (Ours)} & \textbf{65.59} & \textbf{83.11} & \textbf{89.21} \\
					\thickhline
				\end{tabular}
			}
		\end{minipage}
		\hspace{0.45cm}
		\quad
		\begin{minipage}[t]{0.45\linewidth}
			\centering
			\begin{minipage}[t]{1.0\linewidth}
				\centering
				\small
				\makeatletter\def\@captype{table}\makeatother\caption{Comparison with SOTA methods on RSTPReid.} 
				\centering
				\resizebox{1\linewidth}{!}{
					\renewcommand\arraystretch{1} 
					\begin{tabular}[h!]{p{3cm}|m{1.2cm}<{\centering}m{1.2cm}<{\centering}m{1.2cm}<{\centering}} 
						\thickhline
						\textbf{Method} & \textbf{R1} & \textbf{R5} & \textbf{R10}  \\
						\hline
						DSSL \cite{zhu2021dssl} &	 32.43 &55.08 &63.19\\
						\hline
						Baseline (Ours) &	37.40	&60.90 & 70.80 \\ 
						\textbf{IVT (Ours)} & \textbf{46.70 }& \textbf{70.00} &	\textbf{78.80} \\
						\thickhline
					\end{tabular}\label{tab:sota_RSTP}
				}
			\end{minipage}
			\begin{minipage}[t]{1.0\linewidth}
				\vspace{4.5em} 
				\centering
				\small
				\makeatletter\def\@captype{table}\makeatother\caption{Comparison with SOTA methods on ICFG-PEDES.}
				\centering
				\resizebox{1\linewidth}{!}{
					\renewcommand\arraystretch{1} 
					\begin{tabular}[h!]{p{3cm}|m{1.2cm}<{\centering}m{1.2cm}<{\centering}m{1.2cm}<{\centering}}
						\thickhline
						\textbf{Method} & \textbf{R1} & \textbf{R5} & \textbf{R10}  \\
						\hline
						Dual Path \cite{zheng2020dual}   &	38.99 & 59.44 & 68.41\\
						CMPM+CMPC \cite{zhang2018deep}   &	43.51 & 65.44 & 74.26\\
						MIA \cite{niu2020improving}      &	46.49 & 67.14 & 75.18\\
						SCAN \cite{lee2018stacked}       &	50.05 & 69.65 & 77.21\\
						ViTAA \cite{wang2020vitaa}       &	50.98 & 68.79 & 75.78\\
						SSAN \cite{ding2021semantically} &	54.23 &  72.63 &  79.53 \\
						\hline
						Baseline (Ours) &	44.43	& 63.50 & 71.00 \\ 
						\textbf{IVT (Ours)} & \textbf{56.04} & \textbf{73.60}	& \textbf{80.22}	 \\
						\thickhline
					\end{tabular}\label{tab:sota_ICFG} 
				} 
			\end{minipage}  
		\end{minipage}  
	\end{figure}

	\section{Experiment}
	
	\subsection{Experimental Setup}
	\noindent
	\textbf{Datasets.} 
	We evaluate our approach on three benchmark datasets, \emph{i.e.,} \textbf{CUHK-PEDES} \cite{Shuang2017Person},  \textbf{RSTPReid} \cite{zhu2021dssl}, and \textbf{ICFG-PEDES} \cite{ding2021semantically}.  
	Specifically, CUHK-PEDES \cite{Shuang2017Person} contains 40,206 images of 13,003 persons and 80,440 description sentences. It is splitted into a training set with 34,054 images and 68,126 description sentences, a validation set with 3,078 images and 6,158 description sentences, and a testing set with 3,074 images and 6,156 description sentences. 
	RSTPReid \cite{zhu2021dssl} is collected from MSMT~\cite{wei2018person} and contains 20,505 images of 4,101 persons. Each image has two sentences and each sentence is no shorter than 23 words. More in detail, the training, validation, and testing sets have 3,701, 200, and 200 identities, respectively.
	ICFG-PEDES \cite{ding2021semantically} is also collected from MSMT17~\cite{wei2018person} and contains 54,522 images of 4,102 persons. Each image has one description sentence with an average of 37.2 words. The training and testing subsets contain 34,674 image-text pairs for 3,102 persons, and 19,848 image-text pairs for the remaining 1,000 persons, respectively.
	We also explore pre-training on four image captioning datasets: Conceptual Captions (CC)~\cite{sharma2018conceptual}, SBU Captions~\cite{ordonez2011im2text}, COCO~\cite{lin2014microsoft} and Visual Genome (VG)~\cite{krishna2017visual} datasets. There are about 4M image-text pairs in total.

	\noindent
	\textbf{Evaluation Metric.} 
	The cumulative matching characteristic (CMC) curve is a precision curve that provides recognition precision for each rank. Following previous works, R1, R5, and R10 are reported when compared with state-of-the-art (SOTA) models. The mean average precision (mAP) is the average precision across all queries, which is also reported in ablation studies for future comparison.

	\noindent
	\textbf{Implementation Details.}  
	The proposed framework follows the standard architecture of ViT-Base\cite{dosovitskiy2020image}.The patch size is set as 16 $\times$ 16 and the dimensions of both visual and textual features are 768. The input images are resized to 384 $\times$ 128. We use horizontal flipping and random cropping as data augmenting. At the pre-training stage, we utilize 64 Nvidia Tesla V100 GPUs with FP16 training. At the fine-tuning stage, we employ four V100 GPUs and set the mini-batch size as 28 per GPU. The SGD is used as the optimizer with the weight decay of 1e-4. The learning rate is initialized as 5e-3 with cosine learning rate decay.

	\subsection{Comparison with State-of-the-art Methods}
	In this section, we report our experimental results and compare with other SOTA methods on CUHK-PEDES \cite{Shuang2017Person}, RSTPReid \cite{zhu2021dssl} and ICFG-PEDES~\cite{ding2021semantically}. Note that, the Baseline in Table \ref{tab:sota_CUHK}, Table \ref{tab:sota_RSTP} and Table \ref{tab:sota_ICFG}, denotes the vanilla IVT without pre-training, bidirectional mask modeling (BMM) and multi-level alignment (MLA) components. We employ BERT~\cite{kenton2019bert} to initialize the ``txt'' module and ImageNet pre-trained model from~\cite{dosovitskiy2020image} to initialize other modules.

	\begin{table}[t]
		\begin{minipage}[!t]{0.48\linewidth}
			\centering
			\scriptsize
			\caption{Ablation results of components on CUHK-PEDES. ``Base'' denotes our baseline method. ``Pre'' is short of pre-training.}
			\resizebox{1\linewidth}{!}{ 
				\renewcommand\arraystretch{1.1} 
				\begin{tabular}[h!]{p{0.45cm}<{\centering}|m{0.70cm}<{\centering}m{0.70cm}<{\centering}m{0.74cm}<{\centering}m{0.74cm}<{\centering}|m{0.70cm}<{\centering}m{0.70cm}<{\centering}m{0.70cm}<{\centering}m{0.70cm}<{\centering}} 
					\thickhline
					\textbf{No.}    &\textbf{Base}  & \textbf{Pre} & \textbf{BMM} & \textbf{MLA} & \textbf{R1}    & \textbf{R5}    & \textbf{R10}   & \textbf{mAP}   \\ 
					\hline
					1&       \checkmark      &              &               &               & 55.75 & 75.68 & 84.13 & 53.36 \\
					2&       \checkmark      &\checkmark    &               &               & 60.06 & 78.56 & 85.22 & 56.64 \\
					3&       \checkmark      &              &  \checkmark   &               & 60.43 & 79.55 & 86.19 & 56.65 \\ 
					4&       \checkmark      &              &               &  \checkmark   & 61.00 & 80.60 & 87.23 & 56.88 \\
					\hline
					5&       \checkmark      &  \checkmark  & \checkmark    &               & 62.88	& 81.60	& 87.54	& 59.34 \\
					6&       \checkmark      &  \checkmark  &  & \checkmark                 & 63.87 & 82.67 & 88.42 & 59.52     \\
					7&       \checkmark      &              &  \checkmark   &  \checkmark   & 64.00 & 82.72 & 88.95 & 58.99     \\ 
					8&       \checkmark      &   \checkmark & \checkmark    &  \checkmark   & \textbf{65.59} &	\textbf{83.11}&	\textbf{89.21}&	\textbf{60.66}   \\ 
					\thickhline
				\end{tabular}\label{tab:ablation}
			}
		\end{minipage}
		\quad
		\begin{minipage}[!t]{0.48\linewidth}
			\centering
			\scriptsize
			\caption{The computational efficiency terms of several methods. ``Time'' denotes the retrieval time for testing CUHK-PEDES on a Tesla V100 GPU.} 
			\resizebox{1\linewidth}{!}{
				\renewcommand\arraystretch{1.48} 
				\begin{tabular}[h!]{p{1.8cm}|m{2.2cm}<{\centering}m{1.5cm}<{\raggedleft}m{1.5cm}<{\raggedleft}} 
					\thickhline
					\textbf{Method} & \textbf{Architecture} & \textbf{Para (M)}  & \textbf{Time (s)}  \\ 
					\hline
					ViLT~\cite{kim2021vilt}    &Transformer    &96.50    &103,320\\
					ALBEF~\cite{li2021align}   &Transformer    &209.56   & 12,240\\
					\hline
					NAFS~\cite{gao2021contextual}  &ResNet + BERT   &189.00   &78 \\ 
					SSAN~\cite{ding2021semantically} &ResNet + LSTM   &97.86  &31 \\ 
					TBPS~\cite{han2021text}        &ResNet + BiGRU    &84.83  &26 \\ \hline
					\textbf{IVT}    & Transformer    &166.45   &42 \\
					\thickhline
				\end{tabular}\label{tab:efficiency}
			}
		\end{minipage} 
	\end{table}

	\noindent
	\textbf{Results on CUHK-PEDES.} 
	As shown in Table~\ref{tab:sota_CUHK}, our baseline method achieves 55.75\%, 75.68\%, 84.13\% on R1, R5, and R10, respectively. It already achieves comparable or even better performance compared with many works proposed in recent years, \emph{e.g.,} MIA \cite{niu2020improving}, ViTAA \cite{wang2020vitaa}, CMAAM \cite{aggarwal2020text}. These experiments demonstrate the effectiveness of the unified visual-textual network for text-based person retrieval. In contrast, our proposed IVT obtains 65.59\%, 83.11\%, and 89.20\% on these metrics, which are significantly better than our baseline method. Specifically, these results have improved considerably, \emph{i.e.,} +9.84\%, +7.43\%, +5.07\%, respectively. It should be noted that many recent SOTA algorithms have taken complex operations, \emph{e.g.,} segmentation, attention, or adversarial learning. Even though, our approach outperforms existing SOTA algorithms, \emph{e.g.,} DSSL~\cite{zhu2021dssl} and NAFS \cite{gao2021contextual}, and can also be easily implemented. These results fully validate the effectiveness of our approach for text-based person retrieval.

	\noindent
	\textbf{Results on RSTPReid.} 
	As RSTPReid is newly released, only DSSL~\cite{zhu2021dssl} has reported the results on it. As shown in Table~\ref{tab:sota_RSTP}, DSSL \cite{zhu2021dssl} achieves 32.43\%, 55.08\%, 63.19\% on the R1, R5, R10, respectively. In contrast, the proposed method achieves 46.50\%, 70.20\% and 79.70\%, which exceed the DSSL~\cite{zhu2021dssl} by a large margin, \emph{i.e.,} +14.27\%, +14.92\%, and +15.61\%. It should be noted that our baseline method still exceeds DSSL, which benefits from our unified visual-textual network. Besides, our IVT outperforms the baseline with a large margin. The above experiments fully validate the advantages of our proposed modules.

	\noindent
	\textbf{Results on ICFG-PEDES.} 
	The experimental results on the ICFG-PEDES dataset are reported in Table \ref{tab:sota_ICFG}. We can find that the baseline method achieves 44.43\%, 63.50\%, and 71.00\% on the R1, R5, and R10, respectively. Meanwhile, our proposed IVT achieves 56.04\%, 73.60\%, and 80.22\% on these metrics, which also fully validate the effectiveness of our proposed modules for the TPR task. Compared with other SOTA algorithms, \emph{e.g.,} SSAN~\cite{ding2021semantically}, ViTAA \cite{wang2020vitaa}, our results are also better than them significantly. In contrast, even without complex operations to mine local alignments, IVT still achieves SOTA performance.
	
	In summary, our IVT yields the best performance in terms of all metrics on three benchmark datasets. The superior performance is not only due to the well-designed unified visual-textual network, but also owing to the effective implicit semantic alignment paradigms. We hope our work can bring new insights to the text-based person retrieval community.

	\subsection{Ablation Study}
	To better understand the contributions of each component in our framework, we conduct a comprehensive empirical analysis in this section. Specifically, the results of different components of our framework on the CUHK-PEDES \cite{Shuang2017Person} dataset are shown in Table \ref{tab:ablation}.

	\noindent
	\textbf{Effectiveness of Multi-Level Alignment.} 
	As shown in Table \ref{tab:ablation}, the Baseline achieves 55.75\%, 53.36\% on R1 and mAP, respectively. After introducing the MLA module, the overall performance has been improved to 61.00\% and 56.88\%. The improvements up to +5.25\% and +3.52\%, respectively. The results demonstrate the effectiveness of our proposed MLA strategy. Further analysis shows that MLA enables the model to mine fine-grained matching through sentence, phrase, and word-level alignments, which in turn improves the visual and textual representations.

	\noindent
	\textbf{Effectiveness of Bidirectional Mask Modeling.} 
	The BMM strategy also plays an important role in our framework, as shown in Table \ref{tab:ablation}. By comparing No.1 and No.3, we can find that R1 and mAP have been improved from 55.75\%, 53.36\% to 60.43\%, 56.65\%. The improvements up to +4.68\%, +3.29\% on R1 and mAP, respectively. The experimental results fully validate the effectiveness of the BMM strategy for text-based person retrieval.

	\noindent
	\textbf{Effectiveness of Pre-Training.} 
	Even without pre-training, IVT achieves 64\% on R1 accuracy (see No.7), outperforming current SOTA methods, \emph{e.g.,} NAFS (61.50\%), TBPS (61.65\%). To obtain better-generalized features, we pre-train our model using a large-scale image-text corpus. As illustrated in Table \ref{tab:ablation}, we can find that the overall performance can also be improved significantly. Specifically, the R1 and mAP are improved from 55.75\%, 53.36\% to 60.06\%, 56.64\% with the pre-training (see No.1 and No.2). 
	In addition, it can also be found that pre-training improves the final results by comparing the No.5/No.7 and No.8 in Table \ref{tab:ablation}. Therefore, we can draw the conclusion that pre-training indeed brings more generalized features, which further boost the final matching accuracy.

	\begin{figure*}[t]
		\centering
		\includegraphics[width=\linewidth]{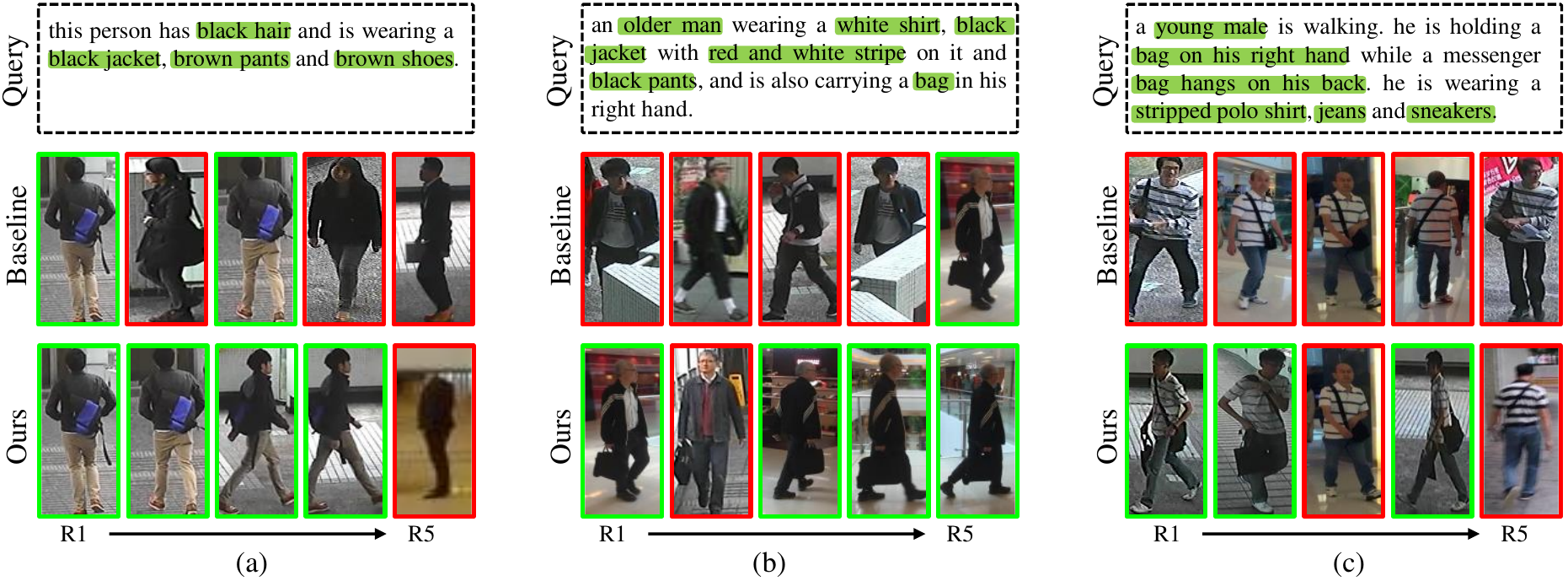} 
		\caption{\textbf{The top-5 ranking results.} The
			\textcolor{green}{green}/\textcolor{red}{red} boxes denote the true/false results.} 
		\label{fig:rank_sota} 
	\end{figure*}
	
	\begin{figure*}[t]
		\centering
		\includegraphics[width=\linewidth]{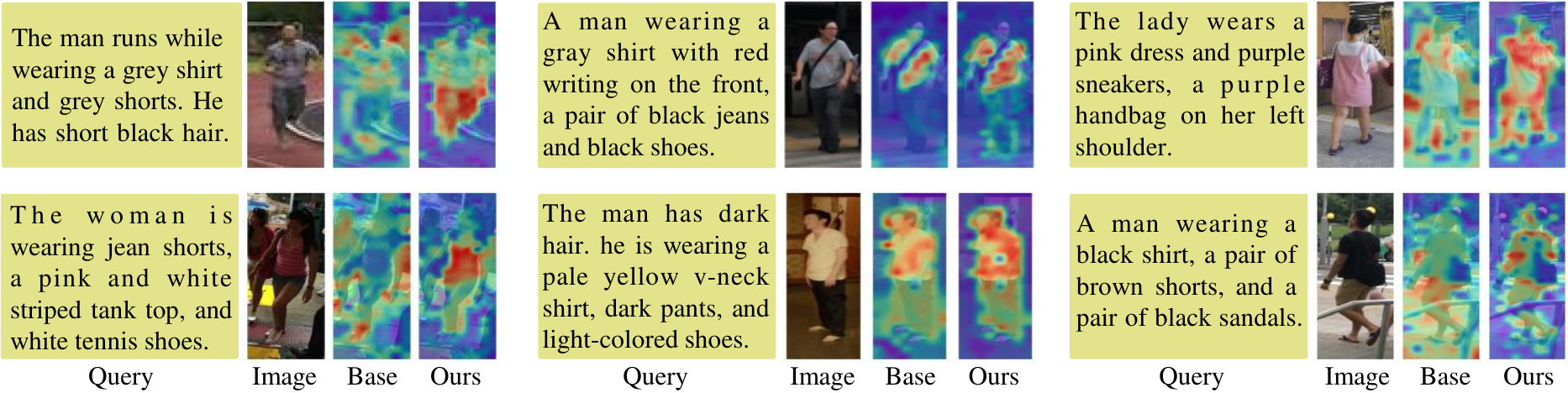} 
		\caption{\textbf{Comparison of heat maps between the baseline method and IVT.} 
		}
		\label{fig:heat_base_ours} 
	\end{figure*}
	
	\begin{wrapfigure}{r}{0.55\textwidth}
		\centering
		\includegraphics[width=0.55\textwidth]{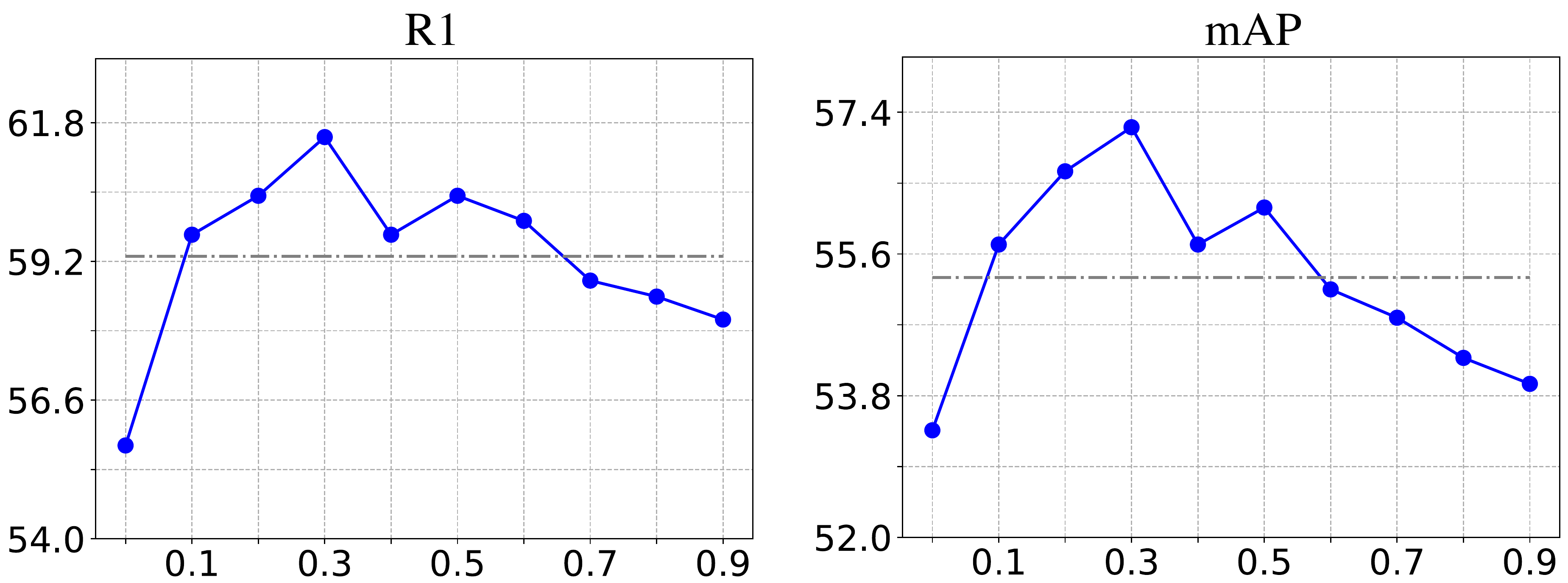} 
		\caption{\textbf{Ablation results of masking ratios on CUHK-PEDES.}} 
		\label{fig:vis_mask} 
	\end{wrapfigure}
	\noindent
	\textbf{Effects of Masking Ratio.}
	The BMM method requires setting the masking ratio. This section studies its effect on final performance. The ablation results are shown in Fig. \ref{fig:vis_mask}. The experiments are conducted on CUHK-PEDES \cite{Shuang2017Person} and only the ``Baseline+BMM'' method is utilized. 
	As shown in Fig. \ref{fig:vis_mask}, the performance has been improved gradually as the masking ratio increases. When the masking ratio is set to zero, which equals the baseline method, the performance is the worst. 
	The R1 and mAP reach their peak values when the ratio is set as 0.3. Then as the masking ratio continues to increase, the performance gradually decreases. This is because the model cannot mine enough semantic alignments with a too large masking ratio, thus reducing the final performance.

	\subsection{Qualitative Results}
	
	\noindent
	\textbf{Top-5 Ranking Results.} 
	As shown in Fig.~\ref{fig:rank_sota}, we give three examples showing the top-5 ranking results. Overall, the retrieved top-5 images show high correlations between the visual attributes and the textual descriptions, even for the false matching results. Compared with the baseline method, our proposed IVT has retrieved more positive samples. This is because it can capture more fine-grained alignments. For example, Fig.~\ref{fig:rank_sota}(b) needs to search for a person with ``white shirt, black jacket with red white stripe on it and black pants''. For the baseline method, the top-1 retrieved image has all these attributes, but ignores other details, \emph{e.g.,} ``older man''. IVT has captured this detail and even captures the attribute ``carrying a bag in his right hand''. Fig.~\ref{fig:rank_sota}(c) shows a difficult case. All the retrieved images have the attributes ``wearing a striped polo shirt, jeans and sneakers'', but all are negative samples for the baseline method. Specifically, the baseline method ignores the description ``holding a bag on his right hand'' while our IVT has captured this detail and retrieved more positive samples.

	\begin{figure*}[t]
		\centering
		\includegraphics[width=\linewidth]{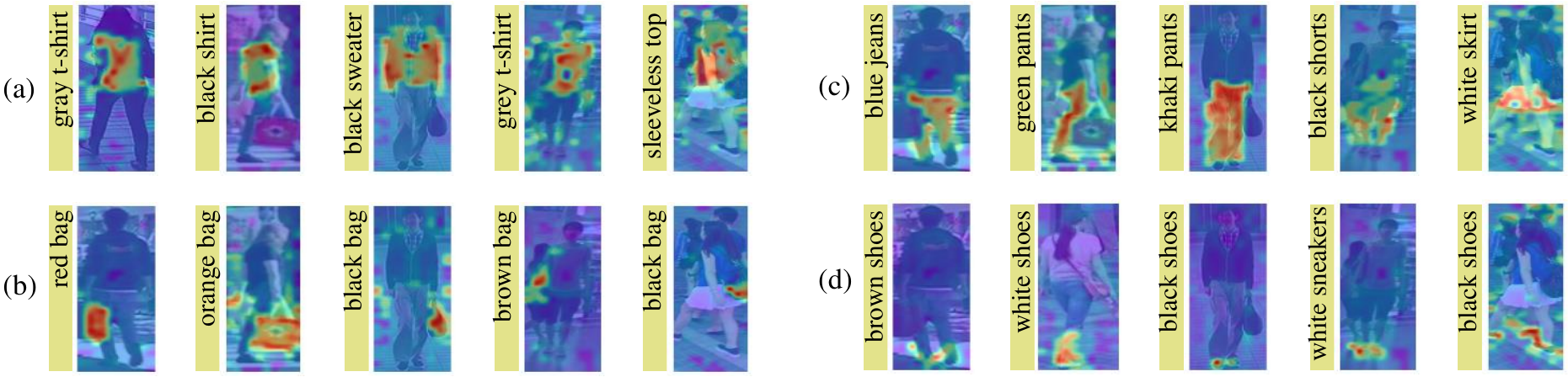} 
		\caption{\textbf{Visualization of part alignment between visual and textual modalities.} (a) Upper Body, (b) Packbag, (c) Lower Body, (d) Shoes. We compute the similarities of word-level text and all image patches. The brighter the part, the higher the similarity.
		} 
		\label{fig:heat_attribute}
	\end{figure*}
	
	\noindent
	\textbf{Visualization of Sentence-Level Heat Map.} 
	To better understand the visual and textual alignment, we give some visualizations of sentence-level heat maps in Fig.~\ref{fig:heat_base_ours}. The heat maps are obtained by visualizing the similarities between textual [CLS] tokens and all visual tokens. The brighter the image is, the more similar it is to the text. In general, textual descriptions can correspond to human bodies, demonstrating that the model has learned the semantic relevance of visual and textual modalities. Compared with the baseline method, IVT could focus more attributes of human bodies described by the text. For example, the man ($1^{st}$ row, $1^{st}$ column) is wearing grey shorts. The baseline method has ignored the attribute, but IVT has captured it. Hence, the proposed IVT can responds to more accurate and diverse person attributes than the baseline method.

	\noindent
	\textbf{Visualization of Local Alignment.} 
	To validate the ability of fine-grained alignment, we further conduct word-level alignment experiments. As shown in Fig.~\ref{fig:heat_attribute}, we give four types of human attributes, \emph{i.e.,} upper body, packbag, lower body, and shoes. Each row in Fig.~\ref{fig:heat_attribute} shares the same attribute. The brighter the area in the image, the more similar it is to the given textual attribute descriptions. As shown in Fig.~\ref{fig:heat_attribute}, our method can recognize not only salient body regions, \emph{e.g.,} clothes and pants, but also some subtle parts, \emph{e.g.,} handbags and shoes. These visualization results show us that our model can focus on exactly the correct body parts, given the word-level attribute description. It indicates that our approach is capable of exploring fine-grained alignments even without explicit visual-textual part alignments. This benefits from the proposed two implicit semantic alignment paradigms, \emph{i.e.,} MLA and BMM. Therefore, we can draw the conclusion that our proposed method indeed achieves \textbf{See Finer} and \textbf{See More} for text-based person retrieval.

	\subsection{Computational Efficiency Analysis}
	In this section, we analyze the parameters and retrieval time at the inference stage.
	As shown in Table \ref{tab:efficiency}, we mainly compare recent methods in the TPR field, \emph{e.g.,} NAFS~\cite{gao2021contextual}, SSAN~\cite{ding2021semantically}, TBPS~\cite{han2021text}, and typical methods in general image-text retrieval, \emph{e.g.,} ViLT~\cite{kim2021vilt}, ALBEF~\cite{li2021align}. Since the parameters of LSTM/BiGRU are less than Transformer, our IVT has more parameters than SSAN and TBPS, but comparable retrieval time. Besides, the performance of these methods would be limited by the text-modeling ability of LSTM. Due to the utilization of Non-local attention, NAFS has 189M parameters and its retrieval time reaches 78s, both exceeding our IVT. Compared with general image-text retrieval methods, \emph{e.g.,} ViLT and ALBEF, our IVT has a significant advantage in retrieval time. Specifically, ViLT needs 103,320 seconds to test CUHK-PEDES, but our IVT only needs 42 seconds. This is because they need to encode all possible image-text pairs, other than just extracting features only once. Overall, our method is competitive enough in terms of both parameters and retrieval efficiency.

	\section{Discussion}
	From the badcases in Fig.~\ref{fig:rank_sota}, we find two characteristics for the TPR task. First, the text description is usually not comprehensive, which corresponds to only part of the visual features. Second, the textual representation tends to ignore subtle features, especially for a relatively long description. By conducting extensive experiments, we got two valuable conclusions: 1) Unified network is effective for the TPR task. It maybe regarded as the backbone network in the future. 
	2) More subtle part alignments should be mined, other than only the salient part pairs. Even without complex operations, the proposed approaches can still mine fine-grained semantic alignments and achieve satisfying performance. We hope they can bring new insights to the TPR community.  
	
	\section{Conclusion}
	This paper proposes to tackle the modality alignment from two perspectives: backbone network and implicit semantic alignment. First, an Implicit Visual-Textual (IVT) framework is introduced for text-based person retrieval. It can learn visual and textual representations using a single network. Benefiting from the architecture, \emph{i.e.,} shared and specific modules, it is possible to guarantee both the retrieval speed and modality interaction. Second, two implicit semantic alignment paradigms, \emph{i.e.,} BMM and MLA, are proposed to explore fine-grained alignment. The two paradigms could see ``finer'' using three-level matchings and see ``more'' by mining more semantic alignments. Extensive experimental results on three public datasets have demonstrated the effectiveness of our proposed IVT framework on text-based person retrieval.

	\section{Broader Impact}
	Text-based person retrieval has many potential applications in surveillance, \emph{e.g.,} finding suspects, lost children, or elderly people. This technology can enhance the safety of the cities we live in. This work demonstrates the effectiveness of unified network and implicit alignments for the TPR task.  
	The potential negative impact lies in that surveillance data about pedestrians may cause privacy breaches. Hence, the data collection process should be consented to by the pedestrian and the data utilization should be regulated.

	\noindent
	\textbf{Acknowledgement.} This work is supported by National Natural Science Foundation of China (NO. 62102205).

	

	
	\clearpage
	%
	%
	\bibliographystyle{splncs04}
	\bibliography{egbib}
\end{document}